\documentclass{article} % For LaTeX2e
\usepackage{iclr2020_conference,times}

\usepackage{amsmath,amsfonts,bm}

\def\eqref#1{equation~\ref{#1}}
\def\1{\bm{1}}

\DeclareMathAlphabet{\mathsfit}{\encodingdefault}{\sfdefault}{m}{sl}
\SetMathAlphabet{\mathsfit}{bold}{\encodingdefault}{\sfdefault}{bx}{n}

\DeclareMathOperator*{\argmax}{arg\,max}

\usepackage{graphicx}

\usepackage{subcaption}
\usepackage{mwe}
\usepackage{bm}
\usepackage{multirow}

\usepackage{hyperref}
\usepackage{url}

\title{How Chaotic Are Recurrent Neural Networks?}

\iclrfinalcopy
\author{Pourya Vakilipourtakalou,\quad Lili Mou\\
Department of Computing Science, University of Alberta\\
Alberta Machine Intelligence Institute (Amii)\\
\url{vakilipo@ualberta.ca}\\
\url{doublepower.mou@gmail.com}
}

\begin{document}
\maketitle
\begin{abstract}
Recurrent neural networks (RNNs) are non-linear dynamic systems. Previous work believes that RNN may suffer from the phenomenon of \textit{chaos}, where the system is sensitive to initial states and unpredictable in the long run. In this paper, however, we perform a systematic empirical analysis, showing that a vanilla or long short term memory (LSTM) RNN does not exhibit chaotic behavior along the training process in real applications such as text generation. Our findings suggest that future work in this direction should address the other side of non-linear dynamics for RNN.
\end{abstract}

\section{Introduction}

Recurrent neural networks (RNNs), e.g., vanilla RNN and Long Short-Term Memory \citep[LSTM,][]{LSTM}, are important architectures for sequential data processing. In natural language processing (NLP), for example, an RNN is used to not only learn the representation of text but also generate a sentence in a certain task~\citep{seq2seq}.

An RNN works in the following way: the RNN keeps a fixed dimensional hidden state. At each time step, it reads in an input signal, updates its state accordingly, and makes a prediction if needed. 
We are particularly interested in RNNs applied to text generation, where at each step, the previously generated word is fed as input and the RNN predicts the next word. 

Mathematically speaking, such RNNs are actually an \textit{iterative map}, or a \textit{map} \citep{mapppa}, from the space of the hidden state to itself. This means that the hidden state is iteratively updated at every step (based on the previous hidden state) by the RNN transition. 

It is noted that, for a typical RNN like vanilla transition and LSTM, the map is \textit{non-linear}, and, a potential known problem of non-linear iterative maps is its \textit{chaotic behavior}~\citep{chaoss}. Usually, a chaotic map is not periodic in the limit of the map being iterated for infinitely many times. Also, a chaotic map prevents an accurate prediction of its state for future steps, in which case we say that the map is attracted to an \textit{strange attractor}. 

Analyzing the chaotic behavior of an RNN becomes a fundamental scientific question for its applications, as chaos behavior might affect how stable an RNN is or how well the RNN performs. 
There have been a few studies on the chaotic behavior of RNNs. 

Both vanilla and LSTM RNNs are shown to be chaotic with certain  parameters~\citep{rnnchaotic,Lchaos}, which are usually assigned by humans in a low dimensional (e.g., 2D) hidden state. Based on such results, a few regularization methods are proposed to make RNNs less chaotic~\citep{Lchaos,regularization}. For simplicity, such studies also ignore the input to the RNN at every step.

In this paper, we analyze the chaos of RNNs (vanilla transition and LSTM) in a more realistic setting in a practical text generation task. We show empirically that---along the entire training process from random weights to a well trained model---the RNN does not exhibit chaotic behavior. Moreover, we experimented the settings of with and without input; empirical results show that, if an RNN is not fed with input, its hidden state is almost always attracted to a single fixed point.

Our unexpected results also imply that, for future work, we should adopt a realistic setting to understand the typical behavior of an RNN in real-world applications.

\section{Formulation}

An \textit{iterative map}, or a \textit{map}, is a function $f: S \rightarrow S$ from one space to the same space. An RNN can be formulated as $\bm h_t=\operatorname{RNN}(\bm h_{t-1}, \bm x_t)$, where $\bm h_t$ is the hidden state for the $t$th step ($\bm h_0$ being the initial hidden state) and $\bm x_t$ is the input. In text generation, for example, we feed in the previously generated word as input, which is usually the most probable word, i.e., $ x_t=\argmax p(x_{t-1}|\bm h_{t-1})$.   This means that such RNN is an iterative map because $\bm h_t=\operatorname{RNN}(\bm h_{t-1}, \argmax p(\cdot|\bm h_{t-1}))$. In previous work where researchers ignore input of RNN, we have $\bm h_t=\operatorname{RNN}(\bm h_{t-1},\bm 0)$, indicating that in this setting the RNN is also an iterative map. These are two scenarios we would explore in this paper, and we refer to them as ``\textit{with input}'' and ``\textit{without input},'' respectively.\footnote{In other usages of RNN (for example, sentence encoding), the input $x_t$ is a word in a given sentence. Such RNN is not well defined as an iterative map, because the function from $\bm h_{t-1}$ to $\bm h_t$ is subject to $x_t$. }

An important notion in the study of iterative maps is the \textit{orbit}, which is the sequence of function values of a map, given an initial input. In the context of an RNN with a given initial hidden state $\bm h_0$, the orbit is $(\bm h_0, \bm h_1, \bm h_2, \cdots)$. 

An orbit may exhibit different behaviors. If $f^k(p)=p$, where $f^k$ means ${f\circ \cdots\circ f}$ for $k$ times, we say the orbit containing $p$ is \textit{periodic} with a period of $k$, or a \textit{period}-$k$ \textit{orbit}. Especially, if $k=1$, the point $p$ is called a \textit{sink}. If an orbit $(\bm h_0,\bm h_1, ....,\bm h_t)$ gets closer to a periodic orbit as $t\rightarrow \infty$, we say the orbit is \textit{asymptotically periodic}.

An orbit may also be \textit{chaotic}. While different researchers do not agree with a common definition~\citep{mapppa, chaoss}, a chaotic orbit usually means that the orbit $(\bm h_0,\bm h_1, \cdots)$ is sensitive to $\bm h_0$, and $(\bm h_0,\bm h_1, \cdots)$ must not be asymptotically period.

In text generation in our experiments, the hidden state of an RNN is usually a high-dimensional space, and it may be difficult to detect periodicity based on RNN's hidden state. We instead approximate it by the RNN's output words. That is, $p(y_t|\bm h_t)= \operatorname{softmax}(W_o \bm h_t+b_o) $, where $W_o$ and $b_o$ are the output parameters. We choose the most  probable word as the predicted word $y_t=\argmax p(y_t|\bm h_t)$. In this case, we call $(y_1, y_2,\cdots )$ an orbit in the output space.

It is noted that the orbit $(y_1, y_2, \cdots )$ being periodic is a necessary condition of $(\bm h_0,\bm h_1, \cdots)$ being periodic. However, we will show that such approximation would be fairly accurate by plotting the latent space with dimensionality reduction. Thus, we will compute the period (if existing) of an RNN map by the period of output words in our experiments. 

For example, if an RNN generates a sequence ``\textit{I like it but , I like it but , $\cdots $}'' for a fairly large number of steps (e.g., 2K--20K), we would empirically estimate the period of this orbit as $5$, and we will show that the hidden states are indeed trapped in $5$ points.

\section{Experiments}

\textbf{Setup.}  
We use the Anna Kerenina textbook as our training corpus.\footnote{\url{https://www.kaggle.com/wanderdust/anna-karenina-book}} The corpus has 400K words, and we split the dataset by 4:1 for training and validation (mainly for early stop). The test phase is to generate sentences following the trained language model, which does not involve input.  

To analyze different types of RNNs, we experimented with the vanilla and the LSTM transitions. The size of the hidden state and embedding size, in both cases, were 300 and 500, respectively. We used Adam as the training algorithm with the initial learning rate 0.001 and other default hyperparameters. %with more than 400 thousand words, as dataset.

Our analysis of how chaotic an RNN is involves running the RNN for a large number of times with different initial hidden states $\bm h_0$. The treatment was slightly different. For the RNN without input, we randomly sampled $\bm h_0$ from a Gaussian distribution centered at $\bm 0$. For RNN with input, we experimented with different initial words $x_1$ fed as input for the first step. In this case, the RNN could be thought of as an iterative map, starting from the second step, with different initial hidden states $\bm h_1$. In total, we had 15K trajectories (different runnings of RNN) in our experiments for each setting.

We detect the period of RNNs by its output words. We ran each trajectory by 2K--20K steps, and then verified the period by another 2K--40K steps. For the setting with longest periods (\autoref{totaltable}), a periodic orbit is verified by roughly 10 times.

%In the case of RNNs with input, for every word in the vocabulary as the initial word and an initial hidden state set to zero, we start generating words and observe the periodicity of the output words and hidden state. Since hidden state is in a 300-dimensional space, in order to visualize its periodicity, we use Principal Component Analysis (PCA) method show a 2-dimensional figure to plot the behavior of the hidden state.

\textbf{Results of RNNs with input.}
We first analyze the behavior of an RNN with input, which is the more realistic usage in applications. We train the model on the training set while monitoring the validation perplexity. We obtained 139 perplexity for vanilla RNN and 108 perplexity for LSTM. We compared our results to previous work~\citep{RNNLM}, and although the corpora are different, we obtained a reasonable perplexity of English.

\begin{figure*}[!t]
        \centering
        %\begin{subfigure}{0.475\textwidth}
        \begin{subfigure}[b]{0.45\textwidth}
            \begin{subfigure}[b]{0.49\textwidth}
            \centering
            \includegraphics[width=\textwidth]{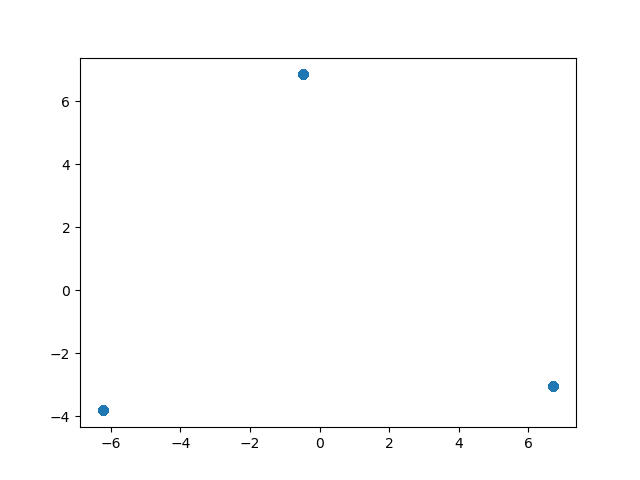}
            %\caption[]%
            {{\small period-3 orbit}}    
            \label{vw4}
            \end{subfigure}
            \hfill
            \begin{subfigure}[b]{0.49\textwidth}
            \centering
            \includegraphics[width=\textwidth]{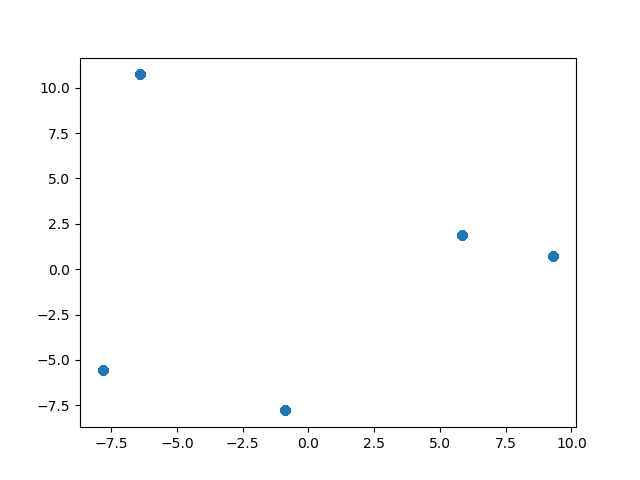}
            %\caption[]%
            {{\small period-5 orbit}}    
            \label{vw1}
            \end{subfigure}
            \vskip\baselineskip
            \begin{subfigure}[b]{0.49\textwidth}
            \centering
            \includegraphics[width=\textwidth]{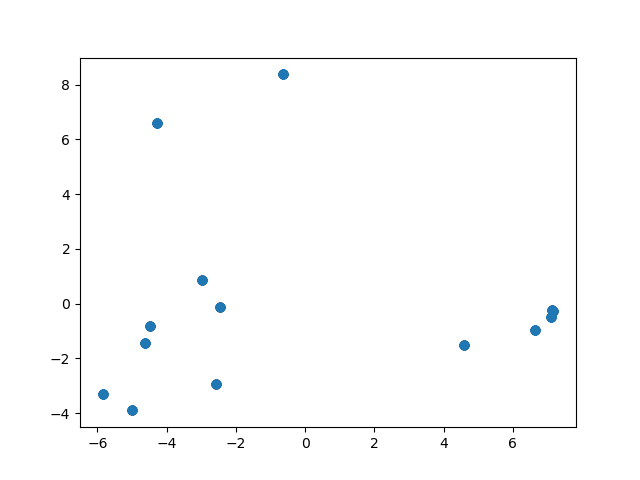}
            %\caption[]%
            {{\small period-14 orbit}}    
            \label{vw2}
            \end{subfigure}
            \hfill
            \begin{subfigure}[b]{0.49\textwidth}
            \centering
            \includegraphics[width=\textwidth]{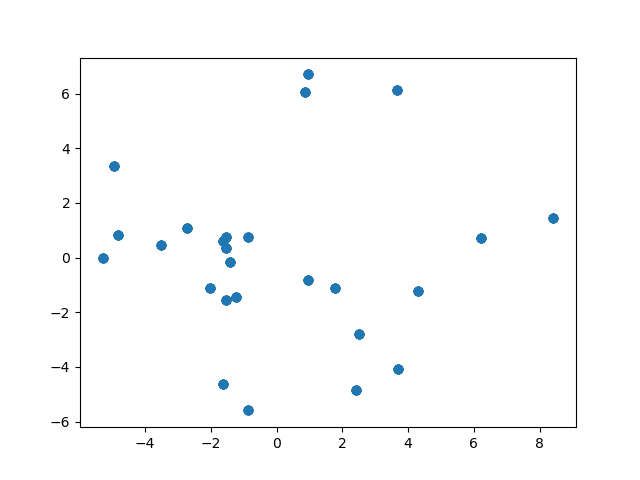}
            %\caption[]%
            {{\small period-26 orbit}}    
            \label{vw3}
            \end{subfigure}
        \caption{Orbits of vanilla RNN}
        \end{subfigure}
        \hfill
        \begin{subfigure}[b]{0.45\textwidth}
            \begin{subfigure}[b]{0.49\textwidth}
            \centering
            \includegraphics[width=\textwidth]{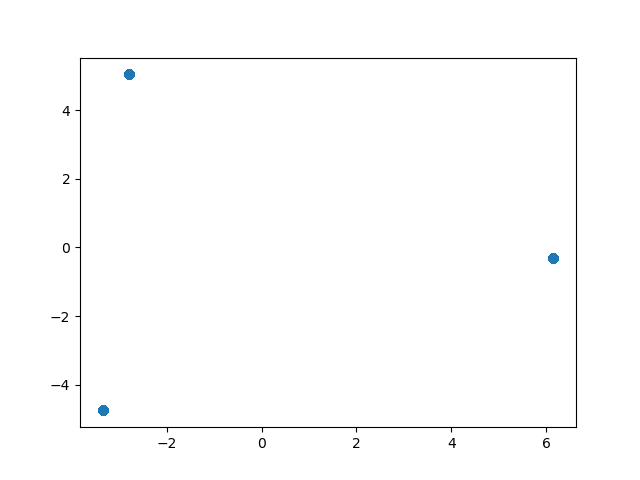}
            %\caption[]%
            {{\small period-3 orbit}}    
            \label{vw5}
            \end{subfigure}
            \hfill
            \begin{subfigure}[b]{0.49\textwidth}
            \centering
            \includegraphics[width=\textwidth]{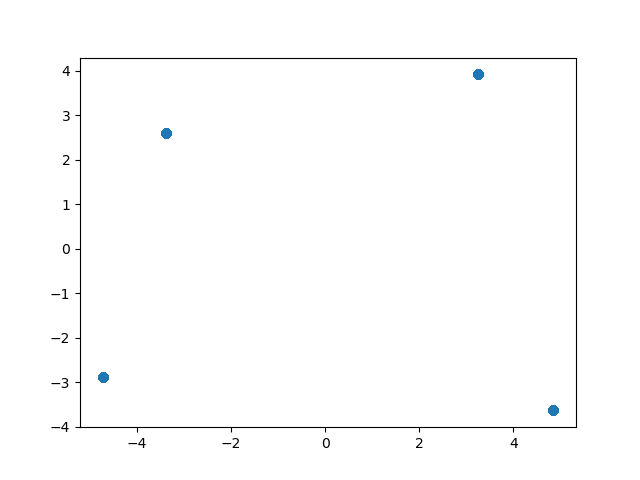}
            %\caption[]%
            {{\small period-4 orbit}}    
            \label{vw6}
            \end{subfigure}
            \vskip\baselineskip
            \begin{subfigure}[b]{0.49\textwidth}
            \centering
            \includegraphics[width=\textwidth]{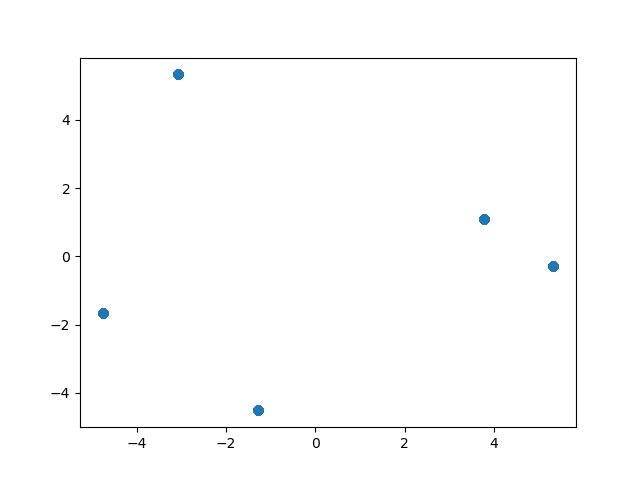}
            %\caption[]%
            {{\small period-5 orbit}}    
            \label{vw7}
            \end{subfigure}
            \hfill
            \begin{subfigure}[b]{0.49\textwidth}
            \centering
            \includegraphics[width=\textwidth]{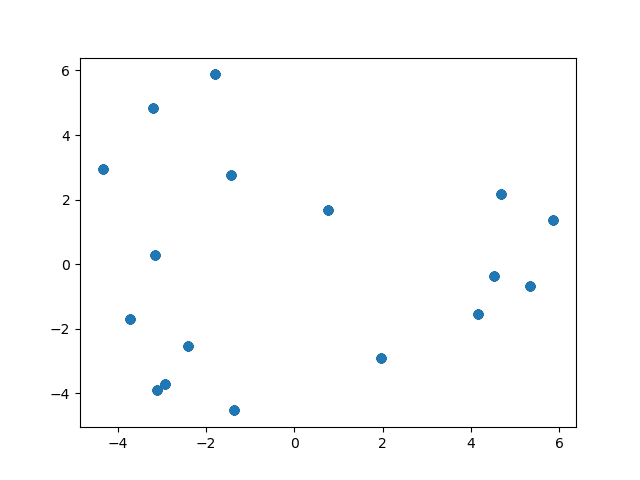}
            %\caption[]%
            {{\small period-17 orbit}}    
            \label{vw8}
            \end{subfigure}
        \caption{Orbits of LSTM}
        \end{subfigure}
        \caption[ ]
        {\small Hidden state of a) vanilla RNN and b) LSTM with input traps in a periodic orbit.} 
        \label{epoc2plots}
\end{figure*}

Unexpectedly, we observe that a well-trained RNN (with input) does not exhibit any chaotic behavior. This is shown in \autoref{epoc2plots}, where we randomly select a few trajectories and plot the high-dimensional hidden state into a 2D space with principle component analysis~\cite[PCA,][]{PRML}. In each plot, we have more than 2K points, but in fact, they are periodic with a length of 3--26.

We are curious if the training process would affect the periodic or chaotic behavior of an RNN. Thus, we quantitatively report with different words the average period of an RNN and the percentage of non-periodic orbits in \autoref{totaltable}. It is seen that all the orbits we obtained are attracted by periodic orbits, showing that they are not chaotic. 

It is also seen that, for both vanilla RNN and LSTM, the average period at the beginning of training is short and then the period becomes longer up to Epoch 20. For vanilla RNN, the average period decreases if we train it more, but we do not observe such decrease for LSTM. Also, the average period of LSTM is significantly longer than that of a vanilla RNN after 20 epochs.

\begin{table}[!t]
\begin{center}
  \begin{tabular}{|c|c|c|c|c|c|c|} \hline
  \multicolumn{2}{|c|}{} & Epoch 0 & Epoch 10 & Epoch 20 & Epoch 30 & Epoch 40 \\  \cline{1-7}
  \multirow{2}{*}{Vanilla RNN} & Average Period & 4 & 39 & 234 & 182 & 101 \\ \cline{2-7}
                                 & Non-Periodic   &0\%& 0\%& 0\% & 0\% & 0\% \\ \cline{1-7} 
  \multirow{2}{*}{LSTM} & Average Period & 3 & 27 & 2408 & 3713 & 4018 \\ \cline{2-7}
                                 & Non-Periodic   &0\%& 0\%& 0\% & 0\% & 0\% \\ \cline{1-7} 
  
\end{tabular}
   \end{center}
   \caption{Vanilla RNN and LSTM trained with input are used to generate words, using the entire words in the vocabulary as input. This table lists the average period of the periodic orbits that the hidden states trap into and also the percentage of the initial words of the vocabulary that perform non-periodic behavior.}
  \label{totaltable}
\end{table}

\textbf{Results of RNNs without input.} 
We also analyzed the behavior of an RNN without input. This is the typical setting of previous work on analyzing RNN's chaotic behavior~\citep{rnnchaotic,Lchaos}. 

\begin{figure*}[!t]
        \centering
        %\begin{subfigure}{0.475\textwidth}
        \begin{subfigure}[b]{0.45\textwidth}
            \begin{subfigure}[b]{0.49\textwidth}
            \centering
            \includegraphics[width=\textwidth]{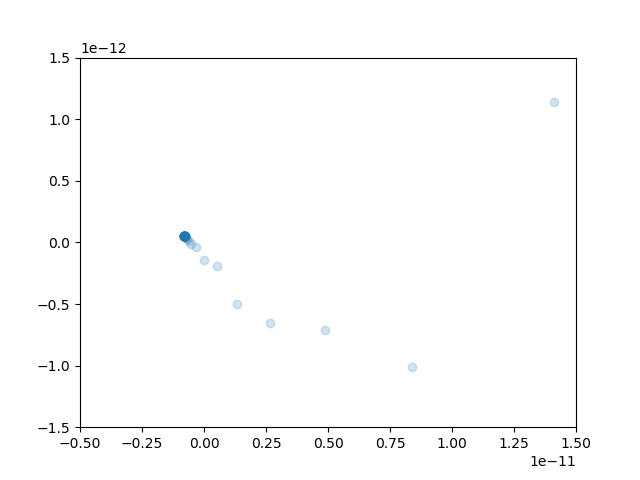}
            %\caption[]%
            {{\small period-1 orbit}}    
            \label{vw11}
            \end{subfigure}
            \hfill
            \begin{subfigure}[b]{0.49\textwidth}
            \centering
            \includegraphics[width=\textwidth]{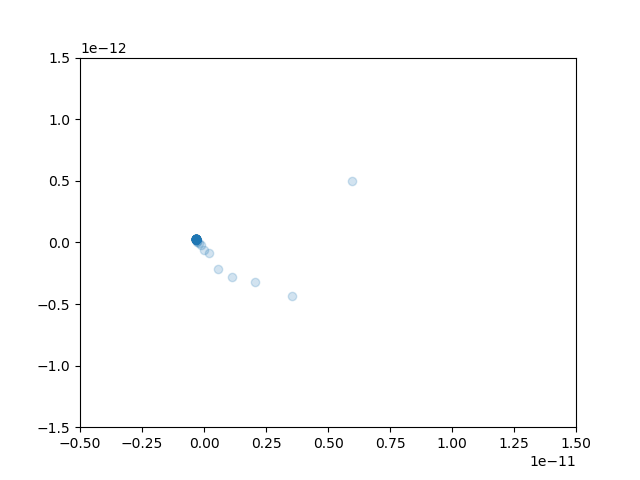}
            %\caption[]%
            {{\small period-1 orbit}}    
            \label{vw12}
            \end{subfigure}
            \vskip\baselineskip
            \begin{subfigure}[b]{0.49\textwidth}
            \centering
            \includegraphics[width=\textwidth]{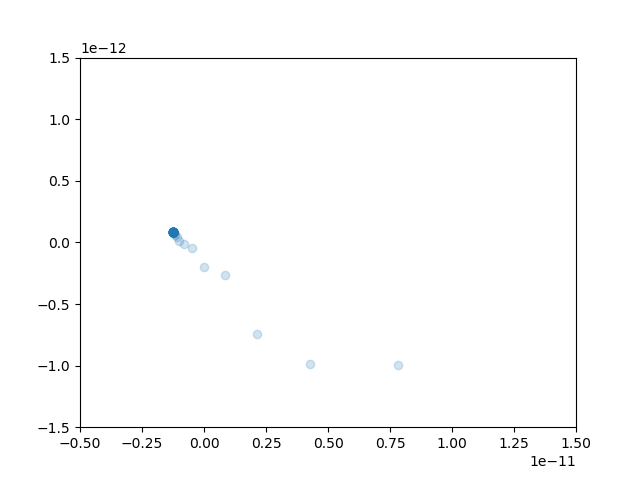}
            %\caption[]%
            {{\small period-1 orbit}}    
            \label{vw13}
            \end{subfigure}
            \hfill
            \begin{subfigure}[b]{0.49\textwidth}
            \centering
            \includegraphics[width=\textwidth]{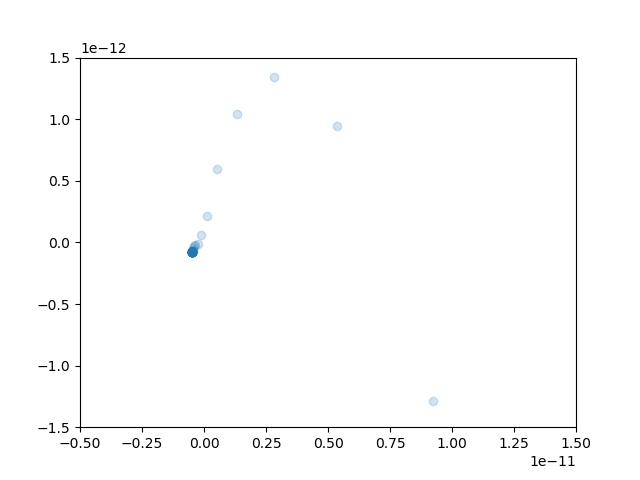}
            %\caption[]%
            {{\small period-1 orbit}}    
            \label{vw14}
            \end{subfigure}
        \caption{Hidden state of vanilla RNN}
        \end{subfigure}
        \hfill
        \begin{subfigure}[b]{0.45\textwidth}
            \begin{subfigure}[b]{0.49\textwidth}
            \centering
            \includegraphics[width=\textwidth]{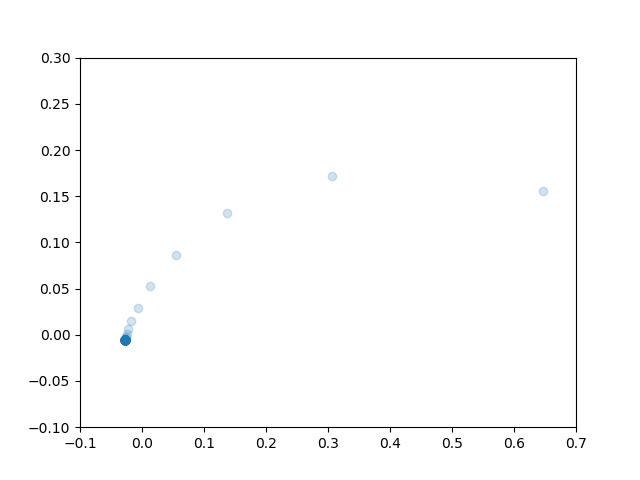}
            %\caption[]%
            {{\small period-1 orbit}}    
            \label{vw21}
            \end{subfigure}
            \hfill
            \begin{subfigure}[b]{0.49\textwidth}
            \centering
            \includegraphics[width=\textwidth]{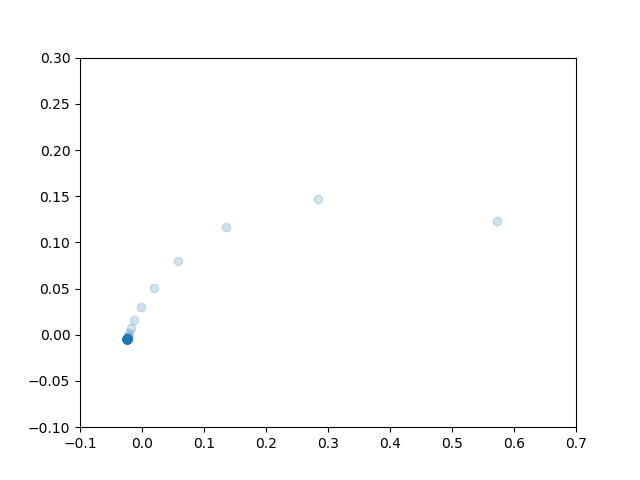}
            %\caption[]%
            {{\small period-1 orbit}}    
            \label{vw22}
            \end{subfigure}
            \vskip\baselineskip
            \begin{subfigure}[b]{0.49\textwidth}
            \centering
            \includegraphics[width=\textwidth]{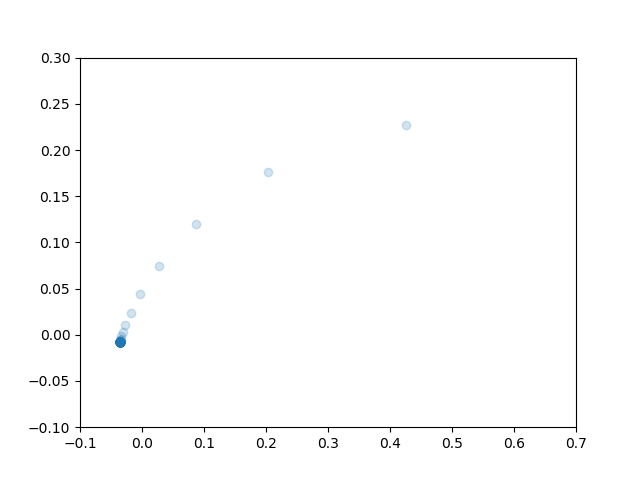}
            %\caption[]%
            {{\small period-1 orbit}}    
            \label{vw23}
            \end{subfigure}
            \hfill
            \begin{subfigure}[b]{0.49\textwidth}
            \centering
            \includegraphics[width=\textwidth]{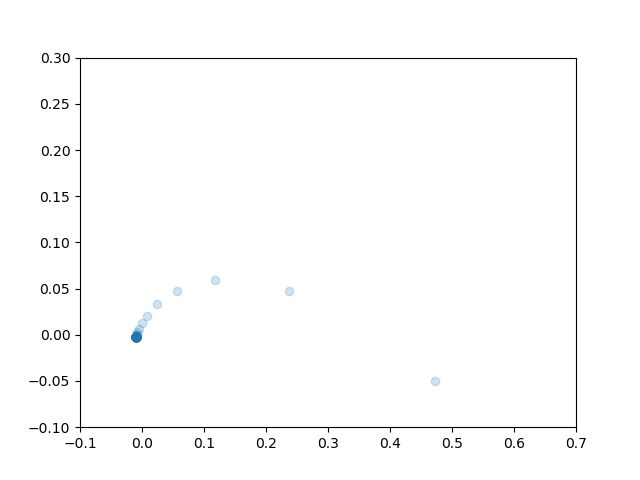}
            %\caption[]%
            {{\small period-1 orbit}}    
            \label{vw24}
            \end{subfigure}
        \caption{Hidden state of LSTM}
        \end{subfigure}
        \caption[ ]
        {\small Hidden state of a) vanilla RNN and b) LSTM without input is attracted to a sink.} 
         \label{figwo}
\end{figure*}

We show the period of such RNNs in \autoref{tablewo}. Again, we do not observe any chaotic behavior for both vanilla RNN and LSTM. More surprisingly, all these orbits are attracted by a sink, i.e., the period is 1. 
This is further confirmed by the PCA plot of RNN's hidden states (\autoref{figwo}). As shown, all the hidden states are monotonically approaching a single point, which is a sink of the RNN. 

In comparison to RNN with input, we see that the period of RNN without input is much shorter, and in fact, all orbits are period-1. Our conjecture is that, if a word is fed to RNN as an input, it is taken from the argmax of the predicted probability in the previous step. Such input word is a discrete token chosen from the vocabulary. This, in turn, may drag the RNN's hidden state and make the period longer.  

It is also noted that a few previous papers report the chaotic behavior of RNN, which seemingly contradicts the observation of our paper. We contacted \citeauthor{} by personal email, and with the help of them, we replicated the chaos with certain RNN weights. Considering the evidence in previous work and our paper, we conclude that vanilla RNN or LSTM could be chaotic with certain weights, but in a realistic setting where the RNN weights are either randomly initialized or well trained, the RNN does not exhibit chaotic behavior.

\begin{table}[!t]
\centering
  
  \begin{tabular}{ | c | c | c |}
    \hline
    Period & Vanilla RNN & LSTM \\ \hline
    1 & 100\% & 100\% \\ \hline
    $>$ 1 & 0\% & 0\% \\ \hline
    non-periodic & 0\% & 0\% \\ \hline
  \end{tabular}
  \caption{Vanilla RNN and LSTM trained without input are used to generate words, with different initialization of the hidden state sampled from a normal distribution between 0 and 1. This table lists the percentage of the initial hidden states that resulted in the specified periodic outputs. }
   
  \label{tablewo}
\end{table}

\section{Conclusion and Future Work}

In this paper, we show that an RNN with either vanilla or LSTM transition is \textit{not} chaotic along the training process in real applications. Our results contradict previous belief that RNNs are chaotic non-linear dynamic systems, although previous work indeed shows that RNNs could be chaotic with certain magic weights. 

Our findings also suggest that, in future work, it is unnecessary to encourage non-chaos for an RNN, because it does not exhibit chaos in real applications. On the contrary, we observe that a better RNN typically exhibits a longer period (e.g., LSTM vs.~vanilla RNN, well-trained vs.~randomly initialized). We conjecture that, in real applications of RNNs, chaos should be encouraged, rather than discouraged, so that the RNN is not attracted to a periodic orbit. 

\section*{Acknowledgments}
This work is partially supported by the Natural Sciences and Engineering Research Council of Canada (NSERC) under grant No.~RGPIN-2020-04465. 
Lili Mou is also supported by AltaML, the Amii
Fellow Program, and the Canadian CIFAR AI Chair Program.

\bibliography{iclr2020_conference}
\bibliographystyle{iclr2020_conference}

\end{document}